\documentclass[10pt,twocolumn,letterpaper]{article}

\usepackage{iccv}
\usepackage{times}
\usepackage{epsfig}
\usepackage{graphicx}
\usepackage{amsmath}
\usepackage{amssymb}

\usepackage{multirow}

\usepackage[breaklinks=true,bookmarks=false]{hyperref}

\iccvfinalcopy % *** Uncomment this line for the final submission

\def\iccvPaperID{****} % *** Enter the ICCV Paper ID here

\ificcvfinal\fi

\begin{document}

%%%%%%%%% TITLE
\title{Causal affect prediction model using a facial image sequence}

\author{Geesung Oh\\
Graduate School of Automotive Engineering\\
Kookmin University, Seoul, Korea\\
{\tt\small gsethan17@kookmin.ac.kr}\\
% For a paper whose authors are all at the same institution,
% omit the following lines up until the closing ``}''.
% Additional authors and addresses can be added with ``\and'',
% just like the second author.
% To save space, use either the email address or home page, not both
\and
Euiseok Jeong\\
Graduate School of Automotive Engineering\\
Kookmin University, Seoul, Korea\\
{\tt\small euiseok\_jeong@kookmin.ac.kr}
\and
Sejoon Lim\\
Department of Automobile and IT Convergence\\
Kookmin University, Seoul, Korea\\
{\tt\small lim@kookmin.ac.kr}
}

\maketitle
% Remove page # from the first page of camera-ready.
\ificcvfinal\thispagestyle{empty}\fi

%%%%%%%%% ABSTRACT
\begin{abstract}
   Among human affective behavior research, facial expression recognition research is improving in performance along with the development of deep learning. However, for improved performance, not only past images but also future images should be used along with corresponding facial images, but there are obstacles to the application of this technique to real-time environments. In this paper, we propose the causal affect prediction network (CAPNet), which uses only past facial images to predict corresponding affective valence and arousal. We train CAPNet to learn causal inference between past images and corresponding affective valence and arousal through supervised learning by pairing the sequence of past images with the current label using the Aff-Wild2 dataset. We show through experiments that the well-trained CAPNet outperforms the baseline of the second challenge of the Affective Behavior Analysis in-the-wild (ABAW2) Competition by predicting affective valence and arousal only with past facial images one-third of a second earlier. Therefore, in real-time application, CAPNet can reliably predict affective valence and arousal only with past data.
   
   The code is publicly available.\footnote{\url{https://github.com/gsethan17/CAPNet_ABAW2021}}
\end{abstract}

%%%%%%%%% BODY TEXT
\section{Introduction}
\label{sec:Intro}
Human affective behavior research is essential for the human-computer interaction (HCI) field and has been established for a long time. However, there are many obstacles for HCI systems used in real-world applications, such as in-the-wild or real-time tasks. To address these problems, Kollias et al. have been hosting the Affective Behavior Analysis in-the-wild (ABAW) Competition, which involves a variety of research activites, for two years \cite{2106.15318, kollias2020analysing, kollias2019face, kollias2021distribution, kollias2019deep, kollias2019expression, kollias2021affect, zafeiriou2017aff}. Most of the top-ranked teams in the first challenge of ABAW (ABAW1) \cite{kollias2020analysing}, held in conjunction with the $15^{th}$ IEEE Conference on Face and Gesture Recognition (FG2020), used convolutional neural networks (CNNs) with single facial images or sequences of such images. In cases where a single image was used, the captured image was inputted to be recognized, and even for teams that used image sequences, past or future images were used along with the image captured at that point \cite{deng2020multitask, kuhnke2020two, li2021technical}. Although these methods perform well with large-scale data in the wild, they encounter limitations when used in real time. Except in cases such as posterior analysis with recorded video, most real-world applications require future decisions only with past data in real time, such as facial expression recognition for interview assistant systems \cite{pampattiwarintelligent} or driver monitoring systems \cite{oh2021drer}.

Hence, we propose a deep learning based network called causal affect prediction network (CAPNet), which enables affect prediction with only past data. CAPNet outputs the predicted affective state in the form of valence and arousal, which are the most popular continuous emotional representations proposed by Russell \cite{russell1980circumplex}. The input data require only a sequence of past facial images, not current or future ones. CAPNet consists of a modular architecture divided into a feature extractor and a causality extractor, which allows one to learn causality well from past facial images. The feature extractor is based on the facial expression recognition (FER) model proposed by Oh et al. \cite{oh2021drer}, which was pretrained using the AffectNet dataset \cite{mollahosseini2017affectnet}. We fine-tuned this FER model with a pair of single images and corresponding labels of the Aff-Wild2 dataset \cite{kollias2018aff} and used the CNN architecture of the fine-tuned FER (FER-Tuned) model as the feature extractor of CAPNet. The causality extractor consists of long short-term memory (LSTM) and fully connected (FC) layers behind the feature extractor. We applied causal inference learning to perform supervised learning by pairing the sequence of past images with the current label using the Aff-Wild2 dataset \cite{kollias2018aff}.

Experiments using the Aff-Wild2 validation dataset \cite{kollias2018aff} showed that CAPNet predicts the affective state well without current or future images. The proposed network achieved lower-quality but nearly similar results compared with those of the FER-Tuned model, which used a single image (current). Nonetheless, CAPNet outperformed the baseline of the second challenge of ABAW (ABAW2) \cite{2106.15318} by generating predictions one-third of a second earlier with only past facial images. Hence, predictions can be generated by CAPNet without having to wait for the current image. The proposed network can predict the affective state from only past data. A considerable advantage of CAPNet is that it can be applied in real time, although a small performance degradation should be expected. Evaluation results obtained from the test set will be notified from ABAW2 \cite{2106.15318}.

The primary contributions of this paper are as follows:
\begin{itemize}
\item We fine-tune the pretrained FER model proposed by Oh et al. \cite{oh2021drer} and propose the performance of the FER-Tuned model using a current facial image as a high baseline for human affective behavior recognition.
\item We propose CAPNet, which uses the order of the input data and the causal structure. CAPNet reliably predicts the affective state early with only past facial images through causal inference learning between the past images current affective state.
\end{itemize}

\section{Related work}
\subsection{ABAW}
Human affective behavior research has been established for a long time and recently produced promising results with developments in deep learning. A contribution to this achievement is the ABAW Competition, hosted by Kollias et al. \cite{kollias2020analysing, 2106.15318}. ABAW consists of three challenges on the same dataset, Aff-Wild2 \cite{kollias2018aff}: dimensional affect recognition (in terms of valence and arousal), categorical affect classification (in terms of the seven basic emotions), and 12 facial action unit detection. Most of the top-ranked teams in ABAW1, which was held in conjunction with FG2020, proposed deep learning based multitask models that output the three challenges at once \cite{deng2020multitask, kuhnke2020two}. For the input data, the corresponding image is basically used, and additional (previous or post) images are used to further leverage temporal information \cite{deng2020multitask, kuhnke2020two, li2021technical}. Kuhnke et al. were ranked a runner-up in the valence-arousal recognition challenge using not only visual data but also audio data as input \cite{kuhnke2020two}. Notably, the top-ranked model in the same challenge used an additional dataset, AFEW-VA \cite{kossaifi2017afew}, along with several methodologies to increase performance. Through ABAW1, human affective behavior recognition performance has been enhanced for in-the-wild environments. However, there are no results considering real-time environments.

\subsection{Causality learning}
Causal learning has been generally overlooked because of the success of machine learning and deep learning. Nonetheless Sch{\"o}lkopf et al. stated that causal inference can improve machine learning and deep learning \cite{scholkopf2021toward}. They said that most of the current success of data-driven models is only the result of large-scale pattern recognition in formalized large data. Hence, causal inference learning is required to solve the problem with generalization outside formalized data. The use of structural causal models, such as modular architectures may enable the development of causality machine learning or deep learning models \cite{scholkopf2021toward}. Entangled recurrent neural networks (E-RNN), proposed by Yoon et al. \cite{yoon2017rnn}, are structural causal models.

Causality learning can also be achieved through sequential input data. It is mainly utilized in generative models such as PixelCNN \cite{van2016pixel} and WaveNet \cite{oord2016wavenet}. PixelCNN \cite{van2016pixel} first defines the order of each pixel in the image. The pixel values are inputted in the defined order, and PixelCNN generates the next pixel. WaveNet \cite{oord2016wavenet} deals with audio data that are made over time. Therefore, the audio data are inputted with a time sequence, and WaveNet generates the next audio data. In these processes, the causality between the previous data and the next data is learned.

Causal inference is also being considered in recent facial expression studies \cite{muller2018causal, shadaydeh2021analyzing}. Therefore, causal learning should also be considered in deep learning based facial expression recognition studies.

\section{Methodologies}
\label{sec:methods}

\begin{figure*}[ht]
\begin{center}
% \fbox{\rule{0pt}{2in} \rule{0.9\linewidth}{0pt}}
   \includegraphics[width=0.8\linewidth]{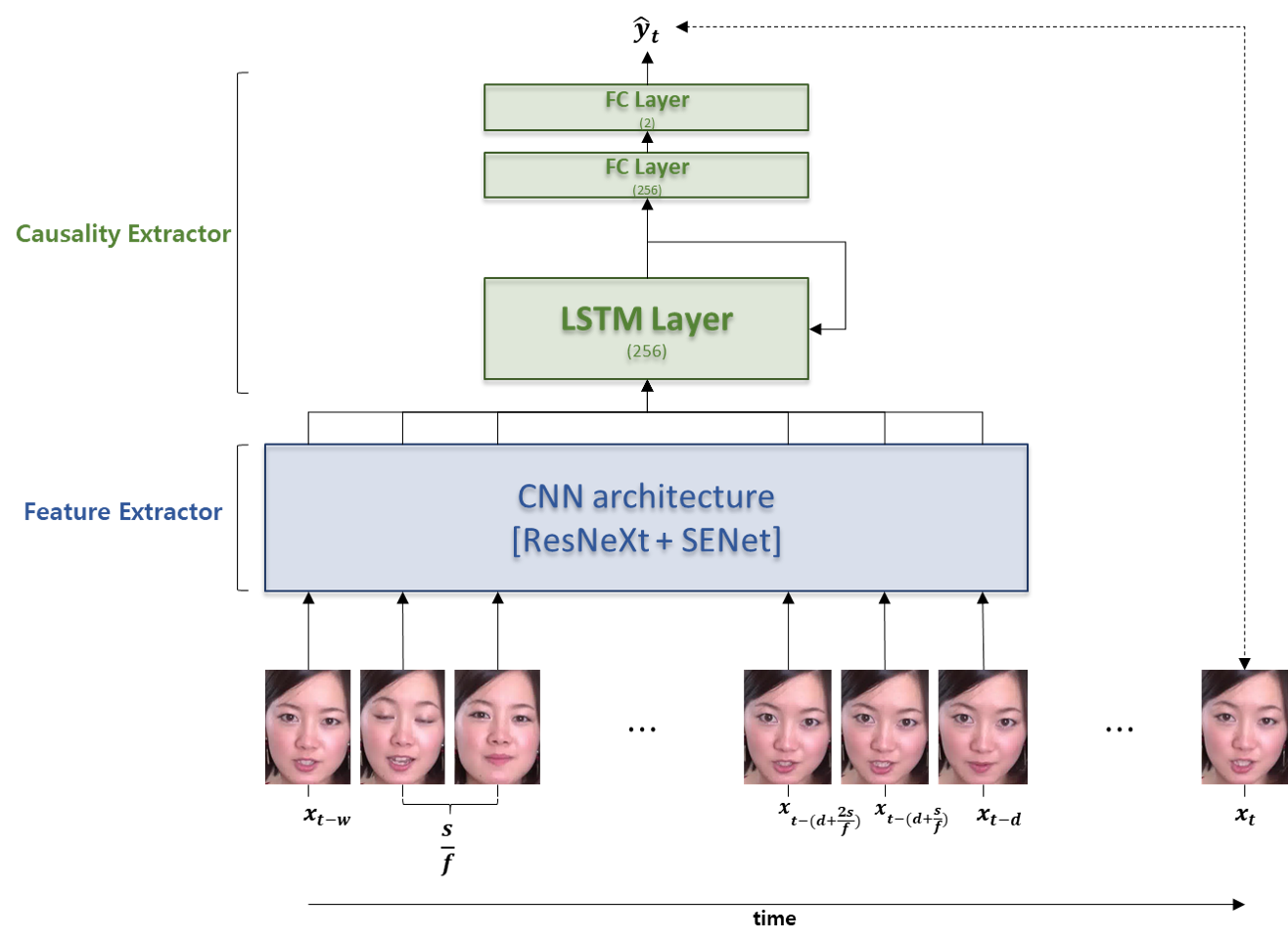}
\end{center}
   \caption{Overall pipeline of CAPNet--the network predicts the current affective state in terms of valence and arousal, $\hat{y}_{t}$, with only past facial images.}
\label{fig:CAPNet}
\end{figure*}

We propose CAPNet, which predicts human affective state indicators with valence and arousal values in past facial images. Through supervised learning using chronologically constructed facial images and corresponding affective labels, CAPNet learns the causal inference between past facial expressions and the current affective state. Multiple past images, past images are created in a single sequence and used as input data. Additional details about the input sequence are presented in Section \ref{sec:input}. Each input image is first extracted into a feature vector through a feature extractor, which is imported from the driver’s real emotion recognizer (DRER) \cite{oh2021drer}. Details about the feature extractor are discussed in Section \ref{sec:fe}. The extracted feature vectors are sequentially fed into a causality extractor, which integrates causal inference from the feature vectors. The way by which the causality extractor integrates the causal information is described in Section \ref{sec:ce}. Figure \ref{fig:CAPNet} shows the overall pipeline of CAPNet.

\subsection{Input sequence}
\label{sec:input}
Ordering of past images, proposed by Oord et el. \cite{van2016pixel, oord2016wavenet}, can be conducted to learn causal inference from past facial images. Even though the order of past images should be determined, most of the images we use are made over time; thus, we only need to organize the order of the input data according to an intrinsic sequence of time steps. Therefore, causal inference is trained on the basis of the sequence of chronological past images and the corresponding labels. CAPNet trains the distribution of valence and arousal over the past facial images of humans under the following conditional distribution:
\begin{equation}
p(y_{t})=p(y_{t}|x_{t-(d + \frac{n}{f})},  n=f \times (w-d), \dots, 2s, s,0)
\end{equation}
where $y_{t}$ is the valence and arousal values of the human at time $t$, $x_{t}$ is the facial expression image of the person captured at time $t$, $d$ is the time duration for prediction, $f$ is the number of images captured per second (frame rate), $w$ is the window size (in seconds), and $s$ is the number of stride images. To predict $y_{t}$, CAPNet needs facial images at intervals of $s$ from time $t-w$ to $t-d$.

\begin{figure*}[ht]
\begin{center}
% \fbox{\rule{0pt}{2in} \rule{0.9\linewidth}{0pt}}
   \includegraphics[width=0.8\linewidth]{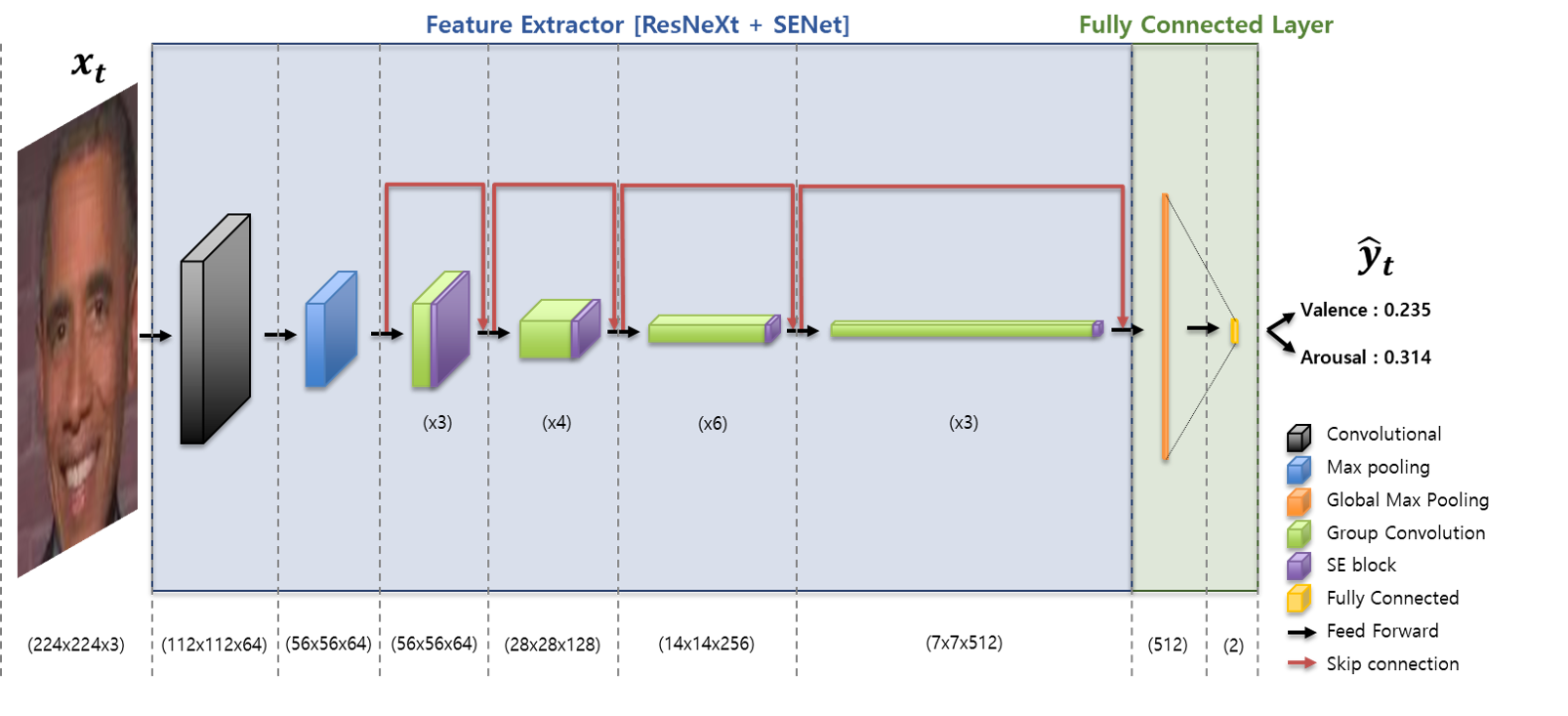}
\end{center}
   \caption{Overall pipeline of the FER model of DRER \cite{oh2021drer}--the FER model recognizes the current affective state in terms of valence and arousal, $\hat{y}_{t}$, with a single (current) facial image. The feature extractor of the FER model is utilized for CAPNet as is.}
\label{fig:FER}
\end{figure*}

\subsection{Feature extractor}
\label{sec:fe}
The CNN architecture we use is based on the FER model of DRER \cite{oh2021drer}, which uses ResNeXt \cite{xie2017aggregated} and SENet \cite{hu2018squeeze} to recognize driver valence and arousal from a single image of the driver's face during driving. The structure of the FER model consists of a feature extractor, which has a CNN structure, and an FC layer, as shown in Figure \ref{fig:FER}. Using the FER model as it is, we can output the valence and arousal of the subject by inputting a single image. Furthermore, the top layer of the FER model (FC layer) can be removed, and the causality extractor can be connected to process a sequence of facial images, as shown in Figure \ref{fig:CAPNet}.

\subsection{Causality extractor}
\label{sec:ce}
Following the proposal of Sch{\"o}lkopf et al. \cite{scholkopf2021toward}, we build the causality extractor as a modular structure so that CAPNet becomes a structural causal model. Since the causality extractor exists independently of the feature extractor, it can be trained individually for the causal inference. The causality extractor consists of one LSTM layer and two FC layers. The LSTM layer lies between the feature extractor and FC layers, and it is based on the LSTM network, proposed by Hochreiter and Schmidhuber \cite{hochreiter1997long}. The LSTM layer outputs a single hidden state for the input sequence of feature vectors generated by the sequence of input images. During the integration of sequential data into the single hidden state, the LSTM layer learn the causal inference between the past facial images and the affective state. The FC layers eventually convert the single hidden state to the predicted affective state. The last FC layer, which has two units and a tanh activation function, outputs a continuous two-dimensional value between $-$1 and 1 representing valence and arousal, respectively.

\section{Experiments}
We studied the models: FER-Tuned (based on the FER model of DRER \cite{oh2021drer}) and the proposed CAPNet. The FER-Tuned model deals with a single image, while CAPNet deals with a sequence of images. 
FER-Tuned was the FER model fine-tuned using the Aff-Wild2 dataset \cite{kollias2018aff}, and the FER model was pretrained with AffectNet \cite{oh2021drer,mollahosseini2017affectnet}, one of the largest datasets composed of single images. CAPNet obtains the feature extractor from the FER-Tuned model, which was trained in the previous step, followed by the causality extractor connected, and trained by the causality extractor over the Aff-Wild2 dataset \cite{kollias2018aff}. In both models, training was conducted only with the training and validation sets of Aff-Wild2 dataset \cite{kollias2018aff}.

\subsection{Data preprocessing}
We used cropped, frame-specific facial images, and the valence-arousal label of each image was provided with 564 videos on the Aff-Wild2 dataset \cite{kollias2018aff}. Except when the cropped images contained detection errors, we used 1,573,844 training sets and 334,865 validation sets for for FER-Tuned model training. These data pairs can be represented by Equation \ref{equ:data}, where $d$ and $n$ are 0.

For CAPNet training, the image sequence is paired with sampling images and labeled as follows:
\begin{equation}
\label{equ:data}
([x_{t-(d + \frac{n}{f})}], y_{t})
\end{equation}
where the set of $x$ contains the cropped past facial images when $n$ is intervals of $s$ from $f \times (w-d)$ to 0 (i.e. $n=f \times (w-d), \dots, 2s, s,0$), $y_{t}$ is the valence-arousal label at time $t$, $d$ is the time duration for prediction, $f$ is the video frame rate, $w$ is the window size (in seconds), and $s$ is the number of stride images. If the cropped facial image does not exist at the $t-(d + \frac{n}{f})$ point, the image will be sampled while navigating sequentially up to the $t-(d + \frac{n+s}{f})+1$ point except $n$ is $f \times (w-d)$. All the videos provided by Aff-Wild2 \cite{kollias2018aff} have a frame rate of $30$ fps; for our experiments, we set $d$ to one-third of a second, $w$ to $3$ s, $s$ to $10$ images. In this case, except when the value of $x$ is insufficient or $y_{t}$ is invalid, we used 1,512,258 training sets and 322,643 validation sets for CAPNet training.

\begin{table*}
\begin{center}
\begin{tabular}{ccccccc}
\hline
\multirow{2}{*}{Model} & \multirow{2}{*}{Input} & \multirow{2}{*}{\begin{tabular}[c]{@{}c@{}}window size\\ (seconds)\end{tabular}} & \multicolumn{3}{c}{CCC}  & \multirow{2}{*}{Remark} \\ \cline{4-6}
\multicolumn{3}{c}{}   & Valence & Arousal & Mean \\
\hline
\hline
VGG-FACE \cite{2106.15318}    & single    & - & 0.23      & 0.21      & 0.22      & baseline   \\

FER-Tuned                & single    & - & 0.520   & 0.516    & 0.518    & \multirow{4}{*}{ours}      \\
\multirow{3}{*}{CAPNet} & \multirow{3}{*}{sequence} & 1 & 0.476    & 0.478    & 0.477\\
                       && 2                      & 0.498    & \textbf{0.486}    & 0.492     \\
                       && 3                      & \textbf{0.510}    & 0.483    & \textbf{0.497}    \\
\hline

\label{tab:validation}
\end{tabular}
\end{center}
\caption{Results on validation set with baseline provided by ABAW2 \cite{2106.15318}}
\end{table*}

\subsection{Implementation}
We trained both models with the following parameters: the input image size was $224 \times 224 \times 3$; the mini-batch size was $128$, and the optimizer we used was Adam \cite{kingma2014adam}, with a 0.00001 learning rate. Training was conducted repeatedly and was terminated if there was no improvement of the validation metric in next four consecutive epochs. The metric used was the concordance correlation coefficient (CCC) \cite{lawrence1989concordance} value, same as that proposed in the ABAW2 competition \cite{2106.15318} and the equation is as follows:
\begin{equation}
CCC=\frac{2 k_{\hat{y}y}}{\sigma_{\hat{y}}^{2}+\sigma_{y}^{2}+(\mu_{\hat{y}}-\mu_{y})^{2}}
\end{equation}
where $\hat{y}$ and $y$ are the model output and label, respectively; $\mu$ is each mean value; $\sigma^{2}$ is each variance; and $k$ is the corresponding covariance value. The average of each CCC value for valence and arousal was used as the overall metric. For training, the loss function for backpropagation was set to $1-$CCC. For CAPNet training, additional parameters were needed. The time duration for prediction ($d$) was wet to one-third of a second; the number of stride images ($s$) was set to $10$ images; and the window size ($w$) was set to 1, 2, and 3 s. Additionally, during CAPNet training, we applied dropout to drop the inputs of the LSTM layer and the first FC layer at a rate of $0.2$.

\subsection{Results}
Even though the Aff-Wild2 dataset \cite{kollias2018aff} contains labels for training and validation sets, labels are not provided for test sets \cite{kollias2018aff}. Hence, we validated the performance of our trained models using the validation set, and the evaluation results using the test set will be notified from ABAW2 \cite{2106.15318}.

% \subsubsection{on the validation set}
Table \ref{tab:validation} presents the results of our methods on the validation set, with ABAW2 results as the baseline \cite{2106.15318}. First, all our proposed models achieved superior CCC values compared with the baseline. The CCC values of valence and arousal were not biased. Although the input of the FER-Tuned model was the same single image as that of the baseline, the mean CCC value was more than 2.3 times the baseline. The overall performance of CAPNet was poorer than that of the FER-Tuned model. Nonetheless, the bigger the window size, the better the performance, and the difference between the FER models was not substantial. In the case of CAPNet, the mean and valence CCC values were the highest when the widow size was 3 s and the arousal CCC value was the highest when the window size was 2 s. In comparison with FER-Tuned in terms of the mean CCC value, CAPNet with a window size of 3 s lost only about 4\% due to starting its prediction one-third of a second earlier.

% \subsubsection{on the test set}

\section{Conclusions}
In this paper, we propose CAPNet, which is based on deep learning and predicts valence and arousal with only past facial images. It outperformed the baseline (ABAW2 results) via fine-tuned of the FER-Tuned model, which predicts valence and arousal through the corresponding single facial image. However, the FER-Tuned model still needs the current image, which hinders its use in real-time application. Hence, we modified FER-Tuned model to enable it to learn causal inference between past images and the current affective state. The modified model is CAPNet, and it can predict current or future affective states with only a sequence of past facial images. In other words, CAPNet can predict at the same time the event occurs or before the event occurs, even when it is applied to systems where facial images are inputted in real time. The validation results showed that corresponding valence and arousal are sufficiently predictable through CAPNet with only past images, one-third of a second earlier.
%than the baseline

However, the performance of CAPNet is slightly poorer than that when the current image is directly used. Thus, filling this gap is one of our future works. We are also considering the use of past audio data. Additionally, we plan to apply CAPNet to real-time environments.

% \section{Acknowledgments}
% TBD

{\small
\bibliographystyle{ieee_fullname}
\bibliography{egpaper_final}
}

\end{document}